\pdfoutput=1
\documentclass[11pt]{article}

\usepackage[]{EMNLP2022}

\usepackage{times}
\usepackage{latexsym}

\usepackage[T1]{fontenc}
\usepackage[utf8]{inputenc}

\usepackage{microtype}

\usepackage{amsmath}
\usepackage{algorithmic}
\usepackage{makecell}
\usepackage{multirow}
\usepackage{amsfonts}
\usepackage{bbding}
\usepackage{graphicx} %插入图片的宏包
\usepackage{float} %设置图片浮动位置的宏包
\usepackage{subfigure} %插入多图时用子图显示的宏包
\usepackage{mathtools}
\usepackage{euscript}
\title{Commonsense Knowledge Salience Evaluation with a Benchmark Dataset in E-commerce}

\author{
	Yincen Qu$^{1}$,
	Ningyu Zhang$^{2}$\thanks{\quad Corresponding author.}, 
	Hui Chen$^{1}$,  
	Zelin Dai$^{1}$, 
	Zezhong Xu$^{2}$, \\
	\textbf{Chengming Wang$^{1}$,  
	Xiaoyu Wang$^{1}$,
	Qiang Chen$^{1}$,
	Huajun Chen$^{2}$,} \\
	$^1$Alibaba Group
	$^2$Zhejiang University \& AZFT Joint Lab for Knowledge Engine \\
	\fontsize{11}{10}\selectfont yincen.qyc@alibaba-inc.com, zhangningyu@zju.edu.cn \\
}
\begin{document}
\maketitle
\begin{abstract}
In e-commerce, the salience of commonsense knowledge (CSK) is beneficial for widespread applications such as product search and recommendation. For example, when users search for ``running'' in e-commerce, they would like to find products highly related to running, such as ``running shoes'' rather than ``shoes''. Nevertheless, many existing CSK collections rank statements solely by confidence scores, and there is no information about which ones are salient from a human perspective. In this work, we define the task of supervised salience evaluation, where given a CSK triple, the model is required to learn whether the triple is salient or not. In addition to formulating the new task, we also release a new Benchmark dataset of Salience Evaluation in E-commerce (BSEE) and hope to promote related research on commonsense knowledge salience evaluation. We conduct experiments in the dataset with several representative baseline models. The experimental results show that salience evaluation is a challenging task where models perform poorly on our evaluation set. We further propose a simple but effective approach, PMI-tuning, which shows promise for solving this novel problem\footnote{Code is available in \url{https://github.com/OpenBGBenchmark/OpenBG-CSK}.}.
\end{abstract}

\section{Introduction}
Commonsense knowledge (CSK) reflects our natural understanding of the world and human behaviors, which are shared by all humans.
In e-commerce, previous study \cite{luo2021alicoco2} have  provided large-scale commonsense knowledge in e-commerce.
However, existing CSK collections \cite{speer2017conceptnet,2019comet,zhang2022aser} rank statements solely by confidence scores, and there is no information about which ones are salient from a human perspective.

\begin{figure}[htbp] %H为当前位置，!htb为忽略美学标准，htbp为浮动图形
\centering %图片居中
\includegraphics[scale=0.15]{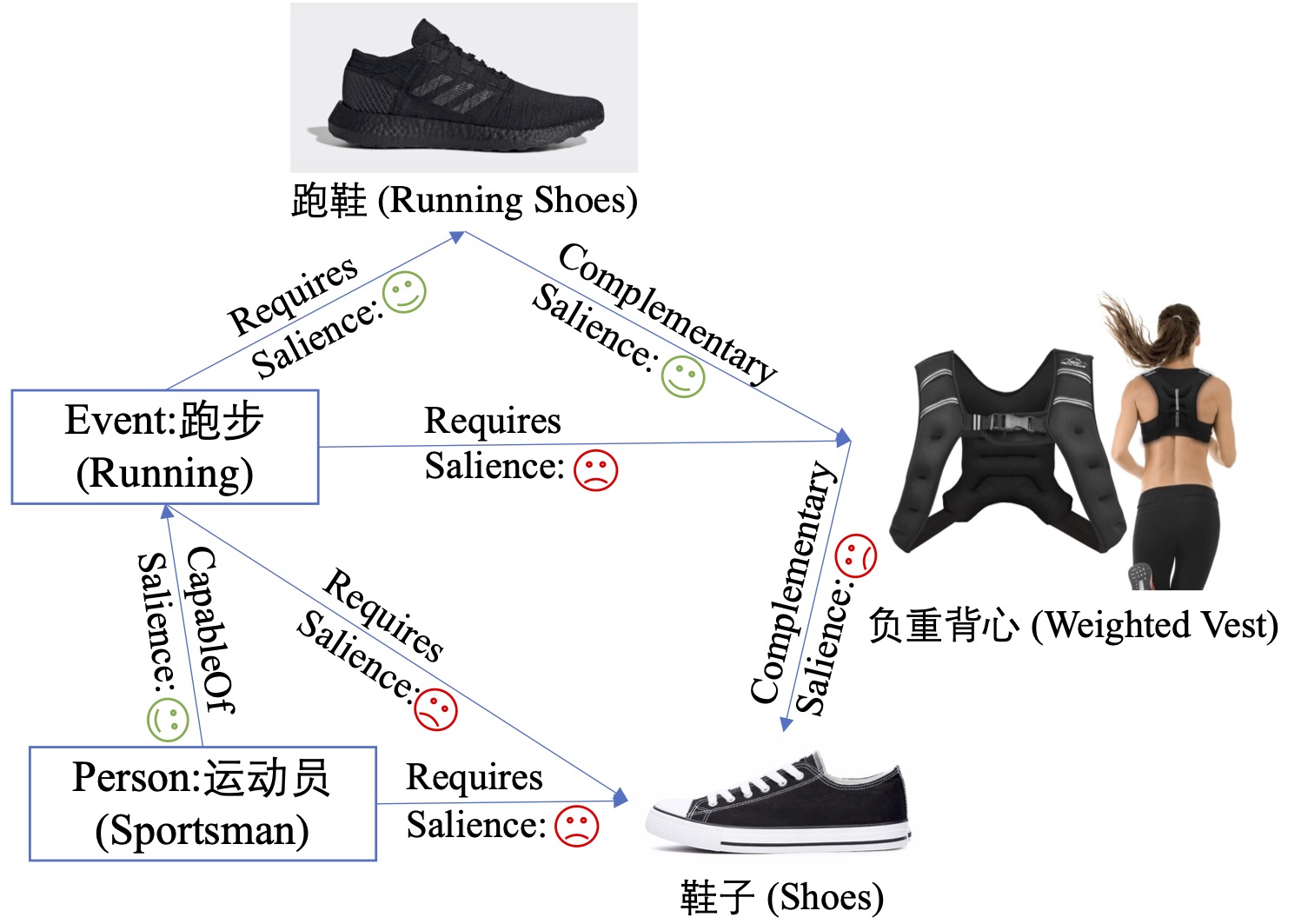} %插入图片，[]中设置图片大小，{}中是图片文件名
\caption{An excerpt from an e-commerce network of commonsense knowledge. Although all the commonsense knowledge is plausible, only the knowledge with a green face is salient, which is beneficial for product search and recommendation. } %最终文档中希望显示的图片标题
\label{figure:illustration} %用于文内引用的标签
\end{figure}

Intuitively, the salience of commonsense knowledge is important and cannot be ignored.
To be specific, salience reflects that a property is characteristic of the concept in the sense that most humans would list it as a key trait of the concept.
For example, lions hunt in packs, bicycles have two wheels, and rap songs have interesting lyrics and beats (but no real melody).
In e-commerce, as shown in Figure \ref{figure:illustration}, ``running'' requires ``running shoes'' is salient (by most people), while ``running'' requires ``shoes'' is not a salient commonsense because for users searching ``running'' in e-commerce, they would like to find products highly related to running, such as running shoes rather than shoes \cite{DBLP:journals/corr/abs-2209-15214}. 
Note that salience captures if the triple is characteristic for the concept, salient common sense can help AI agents understand users' behaviors more explicitly.
Nevertheless, there are few resources and methods regarding commonsense knowledge salience evaluation.

%The problem addressed in this work is to advance CSK collections to a more expressive stage of multi-faceted knowledge.

To this end, we release a commonsense dataset in e-commerce and present a new shared task of commonsense knowledge salience evaluation. 
Unlike previous works, which mostly focus on plausibility, we aim at salience evaluation of every triple in the commonsense knowledge graph. Specifically, we invite annotators to manually annotate the salience for commonsense assertions. As salience is highly subjective, it is hard to annotate directly. Hence we interpret the connotation of salience in terms of necessity and sufficiency and provide annotators with several options standing for the specific grades.

To evaluate the salience evaluation ability of machine learning models, we conduct experiments with several representative baseline models based on our benchmark data. The experimental results show that salience evaluation is a challenging task where models perform poorly on our evaluation set and far below human performance.

Furthermore, we propose a novel PLM-based model, PMI-tuning, to solve this kind of problem.
We introduce the technique of pointwise mutual information (PMI) to estimate the strength of salience. For calculating PMI, we utilize the masked language model to approximate the probabilities of the subject and object. 
We manually define the prompt template for each predicate type and introduce soft prompts based on the idea of prompt tuning in order to probe more focused knowledge from the language model.
\section{Related Works}
\subsection{Commonsense Knowledge Bases}
Commonsense knowledge bases of considerable size have been conducted in several domains, including the generic domain \cite{tandon-etal-2017-webchild,speer2017conceptnet,ATOMIC,mostafazadeh2020glucose} and e-commerce \cite{luo2021alicoco2,10.1145/3447548.3467057}. ConceptNet~\cite{liu2004conceptnet,speer2017conceptnet,speer2013conceptnet} is a well-known knowledge base consisting of triples of entities and relations representing common sense knowledge. 
Alicoco2~\cite{luo2021alicoco2} is the first commonsense knowledge base constructed for e-commerce use. 
Several works~\cite{fang-etal-2021-benchmarking,0.1145discos,he2020role} have aligned well-known knowledge bases and offered evaluation sets to score whether the triple is plausible. However, those commonsense knowledge bases mostly focus on the plausibility of the triples. Recently commonsense knowledge bases such as Quasimodo~\shortcite{romero2019commonsense} and Ascent~\shortcite{chalier2020joint} propose to focus on the salience of assertions. Even so, Ascent does not measure salience quantitatively. Quasimodo merely uses the unsupervised probability formula to rank the salience of the triples.

\subsection{Commonsense Knowledge Scoring Methods}

Commonsense knowledge scoring is widely studied in commonsense knowledge base completion and population tasks. Graph embedding methods~\cite{chen2021relation,abboud2020boxe,lacroix2018canonical,li-etal-2016-commonsense,malaviya2020commonsense} use a graph structure to predict triple plausibility and thus enrich the missing fact in the knowledge base. 
However, graph embedding methods focus on structure learning and ignore contextualized information; they are hard to generalize to unseen triples. Hence, textual encoding approaches~\cite{fang2021discos, yao2019kg,10.1145/3442381.3450043,xie2022discrimination, 2022PKGC} are also applied in such tasks. KG-BERT~\cite{yao2019kg} treats triples in knowledge graphs as textual sequences and fine-tunes a pre-trained language model to predict the plausibility of a triple. 
StAR~\cite{10.1145/3442381.3450043} divides each triple into two asymmetric parts and encodes the two parts into a contextual representation by text encoder. ~\citet{davison2019commonsense} calculates triple plausibility using pointwise mutual information (PMI) between the head entity and tail entity using PLMs. Nevertheless, as an unsupervised method, it suffers from common token bias~\cite{zhao2021calibrate} from pre-trained language models.

A small amount of research work targets salience evaluation. Quasimodo~\cite{romero2019commonsense} uses only unsupervised conditional probability formula to compute salience. 
Dice~\cite{chalier2020joint} ranked assertions jointly along the dimensions of plausibility, typicality, remarkability, and salience using the integer linear programming (ILP) method with soft constraints, which is computationally expensive (10 hours for hyper-parameter optimization on a cluster with 40 cores). Neither work provides a training set for machine learning models to learn salience.

\section{Task Design}
The task requires recognizing, given a commonsense triple, whether it is salient from the human perspective. More concretely, the applied notion of salience is defined as the strength of relatedness between the subject $s$ and the predicate $p$ and the object $o$ in a triple $(s, p, o)$. Statistically speaking, if the co-occurrence of the S and PO is high and the co-occurrence of the S and other PO is low, people would list the PO as a key trait of the S. We say that $(s, p, o)$ is salient if, typically, most humans thinking of $s$ would associate it with $p,o$. This somewhat informal definition is based on (and assumes) common human selectional preference~\cite{resnik1997selectional} as well as commonsense knowledge.

As in other evaluation tasks, the annotators are required to decide whether a triple is salient or not based on the given judgment criteria. Since the definition of salience is marginally vague and hard to annotate, we propose to harness the concepts of sufficiency and necessity to discern salience. The relationship between the salience $\textrm{S}_i$ of instance $i$ and the necessary factor $\textrm{Nec}_i$ and the sufficient factor $\textrm{Suf}_i$ is expressed by the following equation:
\begin{equation}
\begin{aligned}
\textrm{S}_i \!=\!\lambda{\textrm{Suf}_i \!+\!({1\!-\!\lambda}){\textrm{Nec}_i}}\\ 
\end{aligned}
\end{equation}% 
where $\lambda \in [0,1]$ is the parameter that weighs the necessity and sufficiency.

The necessity $\textrm{Nec}$ represents that the subject $s$ is almost the only cause for predicate $p$ and the object $o$ to hold.
While the sufficiency $\textrm{Suf}$ represents that the predicate $p$ and the object $o$ hold for most instances of the subject $s$.
For example, the triple ($\text{running}_s$, $\text{requires}_p$, $\text{shoes}_o$) encodes more sufficiency, since in most situations running requires shoes.
Similarly, ($\text{running}_s$, $\text{requires}_p$, $\text{weight running vest}_o$) encodes more necessity, as running is almost the only cause for requiring weight running vest.
When both necessity and sufficiency are strong, the triple should be salient.
The reasons for leveraging both sufficiency and necessity factors to model the salience is demonstrated in Appendix~\ref{sec:salience}.

\section{Datasets Construction}
To obtain labeled data for training models and as ground truth for evaluation, we construct a dataset in e-commerce. The construction process comprises knowledge acquisition and annotation.
\subsection{Knowledge Acquisition}
To explore the intrinsic relationship between predefined subject and object, we designed a pipeline about relation extraction to acquire implied commonsense knowledge from shopping guidance. Shopping guidance always contains much structured commonsense knowledge in free text. For instance, shopping guidance saying ``The DC motor hair dryer adopts unique intelligent constant temperature control technology to release hair care moisturizing ions...'' implies the common sense of ``hair care requires constant temperature hairdryer''. These texts are obtained by matching the predefined 
subjects and objects. 
With the text found, we can utilize the relationship extraction (RE) model to extract the relationship between the subject and object. The detailed steps for relation extraction are demonstrated in Appendix~\ref{sec:collection}.

\subsection{Annotation Process}
After collecting candidate commonsense triples, we invite annotators to manually annotate each triple's necessity, sufficiency, and salience. Moreover, we investigate the problem of lexical cues and add adversarial examples to avoid the model learning spurious correlations and shallow shortcuts.

\begin{table*}
\centering
\begin{tabular}{lccccc}
\hline
\textbf{Subject}& \textbf{Predicate}  & \textbf{Object} & \textbf{Sufficiency}   & \textbf{Necessity} & \textbf{Salience}\\
\hline
Bride   & Capable Of   & Wedding & Often True & Often True & Salient   \\
Student & Capable Of  & Drinking Water & Often true & Rarely True & Not Salient \\
Running    & Requires    & Wireless Mouse  & Rarely True & Rarely True  & Not Salient  \\
Running   & Requires  & Running Shoes & Often True  & Often True & Salient  \\
Running & Requires   & Weight Running Vest  & Occasionally True &  Often True & Not Salient   \\
\hline
\end{tabular}
\caption{Some cases of manually-annotated triples. Annotators are asked to choose the options of sufficiency, necessity and salience.}
\label{tab:annotation}
\end{table*}

\begin{table}
\setlength{\tabcolsep}{1.4mm}{
\begin{tabular}{lcc}
\hline
\textbf{Examples} & \textbf{Original} & \textbf{Adversarial} \\
\hline
Subject & Running &   Walking\\
Predicate  & Requires & Requires \\
Object  &Running Shoes  & Running Shoes  \\
Sufficiency   & Often True   & Rarely True \\
Necessity  & Often True   & Occasionally True \\
Salience  & Salient   & Not Salient\\
\hline
\end{tabular}}
\caption{Original and adversarial Examples. For the original triple that contains the high coverage word ``running shoes'', the subject is replaced by other words, and annotators confirm the inverted salience score of the new adversarial example. }
\label{tab:examples}
\end{table}

\subsubsection{Annotating}
Annotators were given the collected triples and were required to identify whether the triple was salient or not. Since salience is somewhat equivocal for annotation, we added sufficiency and necessity as auxiliary criteria. 
For sufficiency, annotators need to decide how likely the commonsense triple holds for all instances of the subject, and for necessity, annotators need to decide how likely the subject is the leading cause for the commonsense triple to hold. To reduce the variations, annotators were asked to rate each triple on a three-point Likert scale on sufficiency and necessity. The three options are Often True, Occasionally True, and Rarely True, corresponding to scores of 1, 0.5, and 0, respectively. Finally, annotators were asked to decide whether the triple is salient or not.

For strict quality control, we labeled 100 triples to get the ground truth. Then we carried out a qualification test to select workers. Each worker needed to rate the triples, and their answers were compared with our annotation. 
Annotators who correctly answered at least 8 out of 10 questions were selected. A grading report with detailed explanations on every triple was sent to all workers afterward to help them fully understand the annotation task.

\subsubsection{Adding Adversarial Examples} 
The main source of spurious cues is the uneven distribution of words over labels, as they offer strong signals of the predicted result in inference time. We used the two metrics defined in ~\cite{niven-kao-2019-probing,branco-etal-2021-shortcutted} to detect lexical cues, namely applicability ($\alpha_k$) and coverage ($\xi_k$) of cue $k$.
These metrics are quantitative measures of how likely the presence of n-grams can be a shortcut of models.

We refer the interested reader to Appendix~\ref{sec:cues} for more details about calculating those metrics. The highest coverage $\xi_k$ of words in the training set are all below 0.01. For those characters and words that rank top 1\% for coverage, adversarial examples are obtained by inverting the label of each example and replacing part of a triple with other words (Table ~\ref{tab:examples}). 
Crowd workers are asked to annotate triples' necessity, sufficiency and salience. The confirmed adversarial examples are then added to the original training set.

\begin{table}
\centering
\setlength{\tabcolsep}{2.5mm}{
\begin{tabular}{cccccc}
\hline
Split & \#Train & \#Valid & \#Test   \\
\hline
Random Split & 20,809 & 5000 & 5000   \\
Concept Split  & 21,763 & 4523 & 4523 \\
\hline
\end{tabular}}
\caption{Statistics of the dataset, where \#Ent and \#Rel denote the number of entities and relations. \#Train, \#Valid and \#Test denote the number of triples in the training, validation and test sets, respectively.}
\label{tab:corpus}
\end{table}

\subsection{Settings}

We perform two types of split strategy in the annotated result: {\tt random split}, {\tt concept split}. Random split uniformly divides the data into a training/development/test set randomly. 
Concept split ensures the concept in the training, development, and test sets are disjoint. The concept split is more in line with the real scenarios, and we empirically find (see Section 6.2) that the concept split is more complex than the random split. So the concept split dataset is the main target of our task. In the concept split test set, assertions were triply judged. 
Assertions that could not be agreed upon between annotators on salience options were filtered out. The inter-annotator agreement was `moderate agreement'~\cite{landis1977measurement}, with Fleiss' Kappa values of 0.47, 0.45, and 0.42 for sufficiency, necessity, and salience, respectively. We found that the primary sources of disagreement among annotators are the annotators' different lifestyles and different perceptions of commodities. For example, (shoe washing, requires, laundry detergent) and (treating acne, requires, face toner). 

For the development and test sets, we provide triples and their annotated salience. For the training set, we offer two variants, namely the {\tt simplified} set and the {\tt original} set. For the simplified set, we provide triples and their annotated salience. For the original set, we provide all the annotated sufficiency, necessity, and salience for every triple. The simplified set is more similar to Knowledge Graph Completion (KGC) task, and the relevant KGC models can be used directly in the simplified dataset.

\subsection{Corpus Statistics}

The statistics of our datasets are listed in Table \ref{tab:corpus}. We also list some cases from manually-annotated data in Table~\ref{tab:annotation}. For our version 1.0 dataset, we have collected more than 20,000 instances containing 10,783 entities and 3 relations, all in Chinese. We plan to release a multilingual dataset with more samples in the future.

\begin{figure*}[htbp] %H为当前位置，!htb为忽略美学标准，htbp为浮动图形
\centering %图片居中
\includegraphics[scale=0.094]{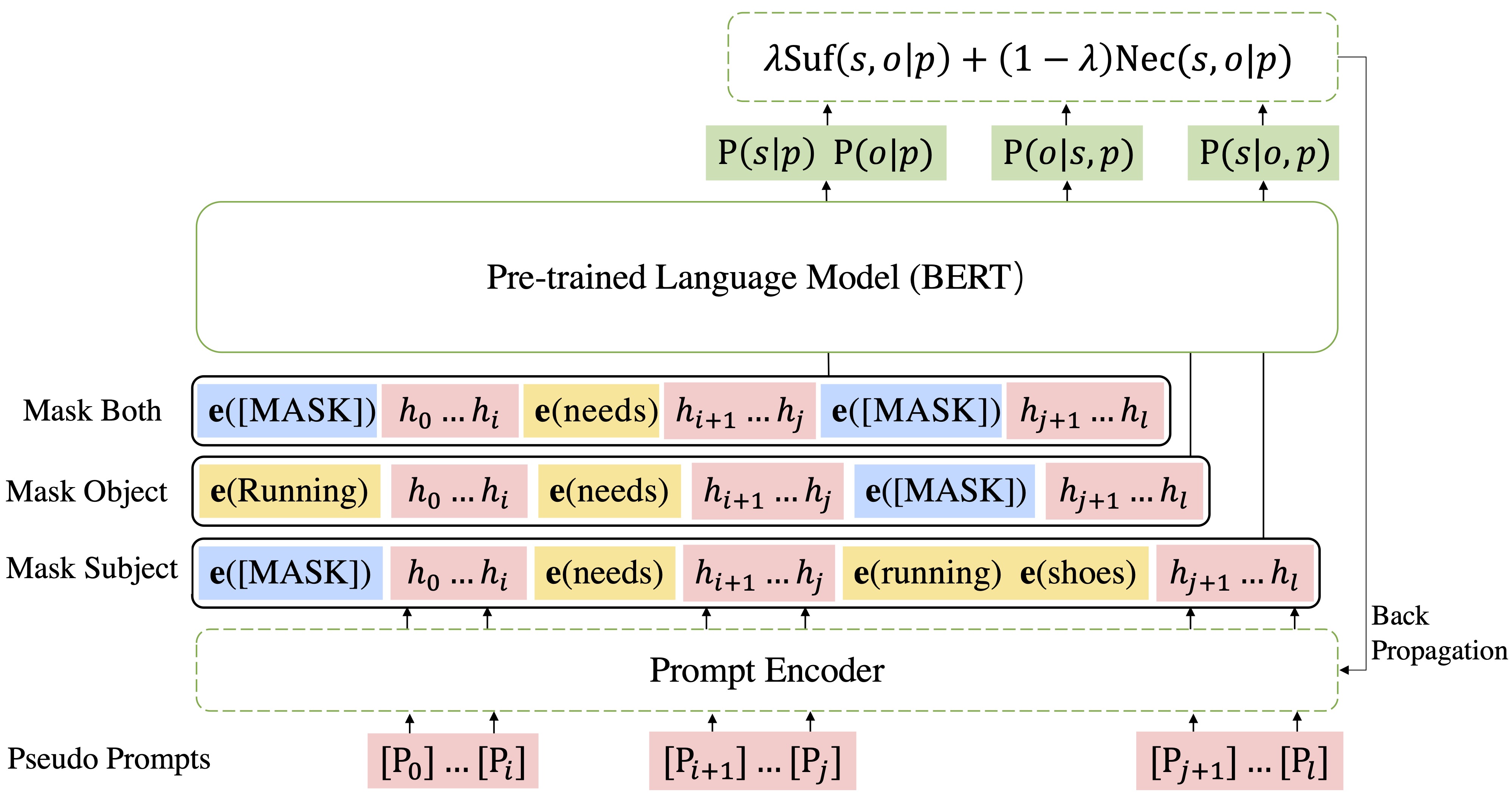} %插入图片，[]中设置图片大小，{}中是图片文件名
\caption{The framework of proposed method PMI-tuning. The input ``Running needs running shoes'' is masked in three ways, including mask head, mask tail, and mask both. Given the context (yellow zone), mask word (blue zone), and the orange zone means prompt token. The pseudo prompt and prompt encoder can be optimized in training while the parameters of the pre-trained language model stay unchanged. } %最终文档中希望显示的图片标题
\label{figure:framework} %用于文内引用的标签
\end{figure*}

\subsection{Baseline Models}
The goal of our benchmark is to determine whether any $(s, p, o)$ triple is salient or not, where the $s$ or $o$ may often be beyond the domain of our training set, and the new entity is unlikely to have a link with existing entities. In this sense, transductive methods based on graph embedding will not be studied here. We present five representative triple classification methods of KGC task merely based on textual encoding, namely BERTSAGE\footnote{\url{https://github.com/HKUST-KnowComp/DISCOS-commonsense}}~\citep{fang2021discos}, KG-BERT\footnote{\url{https://github.com/yao8839836/kg-bert}}~\citep{yao2019kg}, StAR~\citep{10.1145/3442381.3450043}, GenKGC\footnote{\url{https://github.com/zjunlp/PromptKG}}~\citep{xie2022discrimination,DBLP:journals/corr/abs-2210-00305}, PKGC\footnote{\url{https://github.com/THU-KEG/PKGC}}~\citep{2022PKGC}.
The model implementation details will be put in Appendix~\ref{sec:models}.

\section{The Proposed Method}
Driven by the analysis of salience, we propose a simple model to learn salience by utilizing continuous trainable prompt embedding~\citep{liu2021gpt,lester2021power,DBLP:journals/corr/abs-2104-07650}. Figure \ref{figure:framework} presents the framework of the proposed model.

\subsection{Architecture}
Based on the idea that the connotation of salience is necessity and sufficiency, for the triple $(s,p,o)$, the necessity is positively correlated to the conditional probability $\textrm{P}(s|p,o)$ and the sufficiency is positively correlated to the conditional probability $\textrm{P}(o|s,p)$. As conditional probabilities are biased towards highly frequent entities, we utilize marginal probabilities $\textrm{P}(s|p)$ and $\textrm{P}(o|p)$ as the penalty factors. The necessity and sufficiency scores are similar in form to pointwise mutual information (PMI), as shown in the following equations respectively:
\begin{equation}
\begin{aligned}
\textrm{Nec}(s,o|p) &= \textrm{log} \frac{\textrm{P}(s|p,o)}{\textrm{P}^\alpha(s|p)} \\
    & = \textrm{log} \textrm{P}(s|p,o) - \alpha \textrm{log}  \textrm{P}(s|p)
\end{aligned}
\end{equation}
\begin{equation}
\begin{aligned}
\textrm{Suf}(s,o|p) &= \textrm{log} \frac{\textrm{P}(o|s,p)}{\textrm{P}^\alpha(o|p)} \\
    & = \textrm{log} \textrm{P}(o|s,p) - \alpha \textrm{log} \textrm{P}(o|p)
\end{aligned}
\end{equation}
where $\alpha$ is a constant penalty exponent value. We follow ~\citep{luo2016commonsense} to set $\alpha$ to be 0.66, penalizing high-frequency entities.

Noted that the value range of necessity and sufficiency is $(-\infty,+\infty)$, we adopt the form of normalized pointwise mutual information (NPMI)~\citep{bouma2009normalized} to normalize necessity and sufficiency to [-1, 1], where -1(in the limit) represents never occurring together, 0 represents complete independence, and 1 represents complete co-occurrence. The normalized necessity and sufficiency of $(s,p,o)$ are defined as:
\begin{equation}
\begin{aligned}
\textrm{Nec}(s,o|p)\!=\!\frac{\textrm{log} \textrm{P}(s|p,o)\!-\!\alpha \textrm{log}\textrm{P}(s|p)}{ - \textrm{log} \textrm{P}(s|p,o)\!-\!\alpha \textrm{log} \textrm{P}(o|p)}
\end{aligned}
\end{equation}
\begin{equation}
\begin{aligned}
\textrm{Suf}(s,o|p)\!=\!\frac{\textrm{log} \textrm{P}(o|s,p)\!-\!\alpha \textrm{log}\textrm{P}(o|p)}{ - \textrm{log} \textrm{P}(o|s,p)\!-\!\alpha \textrm{log} \textrm{P}(s|p)}
\end{aligned}
\end{equation}

The final salience score encoded by triple $(s,p,o)$ combines $\textrm{Nec}(s,o|p)$ with ${\textrm{Suf}(s,o|p)}$ and is defined as follows:
\begin{equation}
\begin{aligned}
\textrm{S}(s,o|p) \!=\!\lambda{\textrm{Suf}(s,o|p)}\!+\!({1\!-\!\lambda}){\textrm{Nec}(s,o|p)}\\ 
\end{aligned}
\end{equation}% 
where we set $\lambda$ as a learnable variable with a certain initialized value.

To estimate the probabilities, we used the masked language model (MLM), where sentences with masked words constitute model input. To better capture the underlying knowledge in MLM, we organize the masked words, triple, and prompts into different templates. Motivated by LAMA~\cite{petroni2019language}, for every predicate $p$, we manually design a hard template to represent the semantics of associated triples. For example, the hard template for predicate \textit{require} is ``[X] requires [Y].''. By replacing [X] and [Y] with specific subjects and objects, we can obtain the preliminary template. 
Besides, soft prompts are also added to form the final templates. The soft prompts are trainable embedding tensors $[\textrm{P}_0],...,[\textrm{P}_l]$, where $l$ is the predefined prompt length. They are inserted in different positions of the template to make the final sentence more expressive. To compute the marginal probabilities $\textrm{P}(s|p)$ and $\textrm{P}(o|p)$ and conditional probability $\textrm{P}(s|p,o)$ and $\textrm{P}(o|s,p)$, different parts of sentence are masked. Then the three templates become the following:
\begin{itemize}
\item $ \textrm{T}_1\!=\! s [\textrm{P}_{0:i}]p [\textrm{P}_{i+1:j}] \textrm{[MASK]} [\textrm{P}_{j+1:l}]$.
\item $ \textrm{T}_2\!=\! \textrm{[MASK]}[\textrm{P}_{0:i}]p[\textrm{P}_{i+1:j}] o[\textrm{P}_{j+1:l}]$.
\item $ \textrm{T}_3\!=\! \textrm{[MASK]} [\textrm{P}_{0:i}]p[\textrm{P}_{i+1:j}]\textrm{[MASK]}[\textrm{P}_{j+1:l}]$.
\end{itemize}

It is still an open problem to properly compute the probability of words in a sentence with MLM. The recently proposed method pseudo-likelihood scores (PLLs)~\citep{salazar-etal-2020-masked, lin-etal-2021-common} become a proxy of probability, which has shown promising results in many downstream NLP applications. It masks each token $w_i$ at a time and uses the remaining context $s_{\verb|\|i}$ to obtain the probability of a word $w_i$ in the sentence. In this way, the probabilities of subject $s$ and object $o$ in the formatted templates are computed as:
\begin{equation}
\textrm{log} \textrm{P}(o|s,p)=\sum_{i=1}^{|o|}\textrm{log} \textrm{P}(w_i|{T_1}_{\verb|\|i})
\end{equation}%
\begin{equation}
\textrm{log} \textrm{P}(s|p,o)=\sum_{i=1}^{|s|}\textrm{log} \textrm{P}(w_i|{T_2}_{\verb|\|i})
\end{equation}%
\begin{equation}
\textrm{log} \textrm{P}(o|p)=\sum_{i=1}^{|o|}\textrm{log} \textrm{P}(w_i|{T_3}_{\verb|\|i})
\end{equation}%
\begin{equation}
\textrm{log} \textrm{P}(s|p)=\sum_{i=1}^{|s|}\textrm{log} \textrm{P}(w_i|{T_3}_{\verb|\|i})
\end{equation}%

Prompts $[\textrm{P}_0],...,[\textrm{P}_l]$ is modeled as a sequence using a prompt encoder to solve the problems of discreteness and association~\cite{liu2021gpt}. 
Specifically, we use a bidirectional long-short-term memory network (LSTM) to model the association between prompts, with a ReLU-activated two-layer multilayer perceptron (MLP) to encourage discreteness. The real input embeddings prompt $\textrm{P}_i$ of the $i$-th prompt token is defined as:
\begin{equation}
    \textrm{P}_i = \textrm{MLP}([\textrm{LSTM}(\textrm{P}_{0:i}):\textrm{LSTM}(\textrm{P}_{i:l})])
\end{equation}% 

\begin{table*}
\centering
\setlength{\tabcolsep}{4mm}{\begin{tabular}{c|ccc|cccc}
\hline
\multirow{2}{*}{\textbf{Models}} &  \multicolumn{3}{c|}{\textbf{Random Split}} &  \multicolumn{3}{c}{\textbf{Concept Split}} \\
& \textbf{F1}& \textbf{Acc.}  & \textbf{AUC}  & \textbf{F1} & \textbf{Acc.}  & \textbf{AUC}    \\
\hline
BERTSAGE~\citep{fang2021discos}   & 73.1 & 74.2 & 75.7   & 54.5  & 60.1  & 67.1    \\
StAR~\citep{10.1145/3442381.3450043}   & 79.4 & 85.2 & 89.7  & 57.1 & 61.4  & 69.2     \\
KG-BERT~\citep{yao2019kg}  & 95.4 & 97.2 & 98.5 & 59.7 & 63.0 & 70.2  \\
GenKGC~\citep{xie2022discrimination}    & \bf{96.4} & \bf{97.7}  & \bf{99.4} & 60.3 & 60.2 & 71.2   \\
PKGC~\citep{2022PKGC}   & 89.7 & 93.0 & 96.5  & 61.2 & 62.9  & 71.8   \\
\hline
\makecell{PMI-tuning (Simplified)}   & 90.1 & 92.3 & 96.2  & {\bf62.6} & \bf{63.3} & {\bf72.9} \\
\makecell{PMI-tuning (Original)}   &  87.4 & 91.1 & 94.8  & {\bf63.4} & {\bf64.1} & {\bf74.3} \\
\hline
\end{tabular}}
\caption{Test results on salience evaluation. The best score of models is in {\bf{bold}}. All metrics are multiplied by 100. }
\label{tab:plain1}
\end{table*}

\subsection{Training and Inference}
For different variants of datasets, we experiment with different objective functions.
\paragraph{Simplified Set.}Based on the conjecture that the distribution of salience score is a Normal distribution rather than a Bernoulli distribution, we use the mean-squared-error loss rather than cross-entropy loss as the objective function:
\begin{align}
    L = \frac{1}{n} \sum_{i=1}^n (\textrm{S}^{\prime}_i-\textrm{S}_i)^2
\end{align}% 
where $\textrm{S}^{\prime}_i \in \{0,1\}$ is the annotated label of instance $i$ in training set. 
1 denotes the triple is salient, and 0 denotes the triple is not salient. $\textrm{S}_i$ is the predict value of instance $i$.

\paragraph{Original Set.}To better leverage the annotated data, we introduce a variant of circle loss \cite{sun2020circle} to make our model learn the difference between training instances without deliberately constructing pairs of data. 
Since the salience score can be learned through learning comparison of assertions, the pair-wise loss is more reasonable than learning a salient or non-salient label since the annotated labels do not equal the real salience score.
Besides the salience score, we also take necessity and sufficiency into consideration, respectively, as they measure different aspects of salience. So the loss function is formulated as follows:
\begin{equation}
\begin{aligned}
&L = \textrm{log} [1+\sum_{\textrm{S}^{\prime}_i>\textrm{S}^{\prime}_j} \textrm{exp}(\textrm{S}_j-\textrm{S}_i)] \\
&+\gamma \textrm{log} [1+\sum_{\textrm{Nec}^{\prime}_i>\textrm{Nec}^{\prime}_j} \textrm{exp}(\textrm{Nec}_j-\textrm{Nec}_i)] \\
&+ \gamma \textrm{log} [1+\sum_{\textrm{Suf}^{\prime}_i>\textrm{Suf}^{\prime}_j} \textrm{exp}( \textrm{Suf}_j- \textrm{Suf}_i)] \\
\end{aligned}
\end{equation}% 
where $\textrm{Nec}^{\prime}_i$ is the annotated necessity score, $\textrm{Suf}^{\prime}_i$ is the annotated sufficiency score and $\textrm{S}^{\prime}_i$ is the annotated salience value for assertion $i$. $\textrm{Nec}_i$ is the predicted necessity score, $\textrm{Suf}_i$ is the predicted sufficiency score, and $\textrm{S}_i$ is the predicted salience score for assertion $i$. $\gamma$ is a hyperparameter. 
Every sample is compared with others within the mini-batch, and the magnitude of the predicted scores is optimized based on the comparisons.
% test

\paragraph{Inference.} For inference, we directly use Equation 6 with the trained, prompt embeddings to predict the salience score for the current input. Since the real-valued scores for a triple are outputted, we convert these scores into labels by selecting a decision threshold on the development set to maximize the validation F1 score. The threshold is determined by the bisection method.

\section{Experiments}
\subsection{Experimental Setup}
The model needs to determine whether the triple $(s,p,o)$ is salient or not, which is essentially a triple classification task.
Thus, we use the F1, accuracy, and Area under the ROC Curve (AUC) as the metrics for all the test sets.
With two types of split strategy (i.e., random split and concept split), we evaluate all the models on the test set and tune hyper-parameters in the development set. All baseline models are trained based on the simplified dataset.
Besides baseline models, we conduct experiments of PMI-tuning in the simplified training set and original training set.

The textual encoder of all the models are all base models, i.e. BERT-base~\citep{devlin2018bert} and BART-base~\citep{BART}, from the Transformer library\footnote{\url{https://transformer.huggingface.co/}.}. We use batch size 8 and Adam~\citep{Adam} with an initial learning rate of $1\times10^{-5}$ for all models. For PMI-tuning, the template of the prompt is $s\ [\textrm{P}_{0:2}]\ p \ [\textrm{P}_{3:6}]\ o\ [\textrm{P}_{7:11}]$. For hyperparameters, we set $\gamma=0.1$ and $\lambda=0.5$.

\subsection{Results}
\subsubsection{Main Results}
In Table \ref{tab:plain1}, we give the experimental results of all models. It is noted that PMI-tuning achieves the best score on the concept split. On the random split, GenKGC achieves the best performance. By comparing the performance of the model in random split and concept split side-by-side, we can find that all models perform better on the random split than on the concept split, which illustrates that OOV commonsense salience evaluation is a challenging task. In the concept split dataset, we performed a dependent samples t-test of the experimental results between PMI-tuning (simplified) and PKGC (the second-best result). The calculated t-statistic is 5.19. The p-value is 0.003 (p<0.05), which indicates that we have evidence against the null hypothesis of equal means and that the difference is significant. PMI-tuning (original) achieves better performance than PMI-tuning (simplified) in the concept split, implying that the joint training of sufficiency and necessity is also beneficial for generalization.

From the table, we can see that the cross-encoder models, such as KG-BERT and PKGC, tend to be superior to bi-encoder models, such as StAR in this task. As the graph is very sparse in our dataset, models that employ embedding-based representations are hard to get fully trained. It indicates that learning the deeper interactions between words can help the model to better determine whether the triple is salient or not. PMI-tuning, on the other hand, leverages PMI to capture the latent interactions in triples, which is beneficial for achieving better-generalized performance in unseen concepts. As discussed before, PMI is an effective measure to quantify the discrepancy between the probability of subjects and objects, and it is more beneficial for capturing salience than only leveraging text information.

Moreover, we compare our model with Dice~\cite{chalier2020joint} on their annotated pair-wise preference (ppref) test set. Note that we do not have the ground-truth training set; we train PMI-tuning (original) on the results of Dice. 
The experiment details are put in Appendix~\ref{sec:experiment}. The comparison results of salience on the test set of Dice are shown in Table \ref{tab:dice}. As can be seen from the table, despite training on Dice's results, with prior knowledge from the language model, our model consistently outperforms Dice on all the knowledge bases. Besides, Dice takes around 10-14 hours for optimization on a cluster with 40 cores; it takes us only 1 hour to train PMI-tuning (original) in a Tesla P100 GPU card.

\begin{table}
\centering
\setlength{\tabcolsep}{1mm}{
\begin{tabular}{cccc}
\hline
&\textbf{ConceptNet} & \textbf{TupleKB} & \textbf{Quasimodo}   \\
\hline
Baseline & 0.54 & 0.59 & 0.53   \\
Dice & 0.65 & 0.61 &0.63 \\
PMI-tuning & {\bf0.69} & {\bf0.62} & {\bf0.66}   \\
\hline
\end{tabular}}
\caption{Comparison results on the test set of Dice~\cite{chalier2020joint}.}
\label{tab:dice}
\end{table}

\subsubsection{Analysis}
In order to further investigate the effects of components in our model, we conduct more experiments on the concept split. Our analysis mainly focuses on the effects of model components and model performance in the low-resource setting. Due to space limitations, we put more analysis in Appendix ~\ref{sec:Ablation}.

\paragraph{Effect of the Encoders.} We evaluate different PLMs as encoders. Specifically, we choose RoFormer~\citep{su2021roformer}, RoBERTa~\citep{liu2019roberta}, BERT-wwm~\citep{cui2021pre} and MacBERT~\citep{cui2020revisiting}. The experimental results are shown in Table \ref{tab:ablationencode}. As we can see from the table, BERT-wwm achieves the best performance. 

\paragraph{Effect of the Factors.} To study the impact of necessity and sufficiency factors, we utilize different values of $\lambda$ in experiments. The value of $\lambda$ is set to 0 means that only the necessary factor is considered, and it is set to 1, which means that only the sufficient factor is considered. We set $\lambda \in \{0,0.3,0.5,0.7,1\}$ to investigate the importance of necessity and sufficiency. The results are shown in Table \ref{tab:ablation}. From the table, we can see that $\lambda=0.5$ gives the best result, suggesting that necessity is roughly as important as sufficiency.

\begin{figure}[htbp] %H为当前位置，!htb为忽略美学标准，htbp为浮动图形
\centering %图片居中
\includegraphics[scale=0.52]{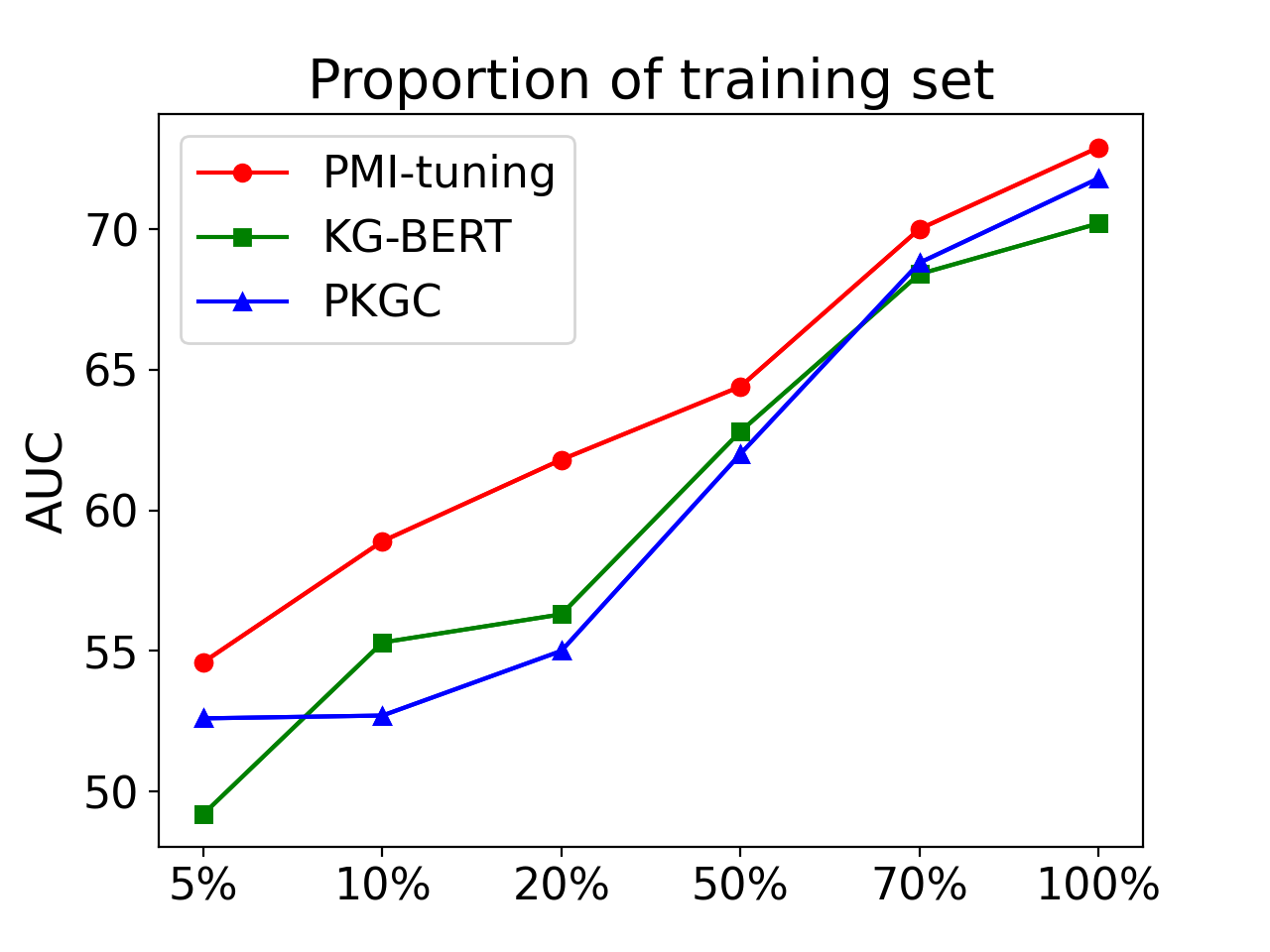} %插入图片，[]中设置图片大小，{}中是图片文件名
\caption{The experimental results on the concept split for a different proportion of training set.} %最终文档中希望显示的图片标题
\label{figure:proportion} %用于文内引用的标签
\end{figure}

\paragraph{Low Resource Setting.} 
Unlike other models referred to in this paper, our model is able to utilize vanilla PLMs to compute reasonable salience without training. Therefore, we conjecture that our model is insensitive to the amount of training data. To validate it, we train models using different proportions of the training set and get the performance.

The experimental results are shown in Figure \ref{figure:proportion}. We notice that when the amount of data used for training decreases, the performance of our model drops less than the performance of both KG-BERT and PKGC. Especially on the 20\% proportion of the training set, PMI-tuning outperforms other models by a large margin. 
This indicates that our model is less sensitive to the amount of training data compared to other models and has the potential to be used for low-resource settings.

\section{Conclusion and Future Work}
%Commonsense knowledge has been widely used in AI applications, such as question answering and reading comprehension. 
We propose a framework of PMI-tuning that could deal with salience scores by probing the latent knowledge distribution in masked language models. Besides, we annotate data for quantitatively evaluating the necessity, sufficiency, and salience of common sense. The experimental results show that evaluating the salience score remains a technical challenge, where most models exploit spurious lexical cues while prompt learning is beneficial to learning the intrinsic salience of a triple. Along with this paper, we publish the annotated dataset and hope that this dataset will facilitate future research on salient commonsense knowledge.

In addition, our future work plans include leveraging this salient CSK for product recommendation and search in E-commerce. 
Moreover, we found a positive correlation exists between Click-Through Rate (CTR) and salience of triple. It indicates the potential of salience to play a greater role in search. We put the detailed verification process and exploratory applications in Appendix ~\ref{sec:appendix}.

\section{Limitations}
There are still many limitations to our dataset and method.

It is a dilemma for us to choose a proper way of annotation.
We noted that a more ideal way of annotation is to give one pivot sample $(s,p)$ (e.g. `running requires:') and rank candidates $o$ with different salience levels (e.g. `water', `running shoes', `shoes', `weighted vest', `cucumber'). In other words, it is more intuitive for human to compare $(s,p,o) > (s,p,\hat{o})$ than determine $(s,p,o)$ is 0 or 1. Nevertheless, in practice, it is hard to collect pair-wise candidates for each pivot sample. 
Moreover, the pair-wise results have the problem that they are not well calibrated to an absolute notion of salience, which brings problems when used to filter non-salient candidates.

Moreover, we do not focus on coverage when constructing the dataset. We only select several e-commerce-related commonsense relations, and other relations are omitted. 
Furthermore, considering that the training and development sets were only annotated once and the annotation was subject to the cognitive biases of different individuals, the results are not entirely accurate.

For the proposed method, we noted that PMI-tuning has low scalability to long text, which is restricted to the max length of 512. 
Moreover, due to mean-squared error loss, PMI-tuning (simplified) requires more training epochs than normal triple classification models. 
We empirically find that the training epoch to reach the stability of other methods, such as KG-BERT, is around 5, but the training epoch to reach the stability of PMI-tuning (simplified) is around 10.

\section*{Acknowledgment}

We want to express gratitude to the anonymous reviewers for their kind comments.
This work is part of the OpenBG Benchmark\footnote{\url{https://github.com/OpenBGBenchmark/OpenBG}} \cite{DBLP:journals/corr/abs-2209-15214}, which is a large-scale open business knowledge graph evaluation benchmark based on OpenBG\footnote{A large-scale open business KG with 2.6 billion triples released by Alibaba Group at \url{https://kg.alibaba.com/}. } to facilitate reproducible, scalable, and multimodal KG research.
Welcome submission in \url{https://tianchi.aliyun.com/dataset/dataDetail?dataId=122271&lang=zh-cn}.

This work was supported by the National Natural Science Foundation of China (No.62206246, U19B2027 and 91846204), Zhejiang Provincial Natural Science Foundation of China (No. LGG22F030011), Ningbo Natural Science Foundation (2021J190), and Yongjiang Talent Introduction Programme (2021A-156-G). 
This work was also supported by Alibaba Group through Alibaba Innovative Research Program.

% Entries for the entire Anthology, followed by custom entries
\bibliography{anthology,custom}
\bibliographystyle{acl_natbib}

\appendix

\section{Salience Definition}
\label{sec:salience}
A fundamental phenomenon of natural language is the Selectional Preferences (SP), where given the word and a dependency relation, human beings have preferences for which words are likely to be connected. For instance, when seeing the verb `sing,' it is highly plausible that its object is `a song.' SP over linguistic relations can reflect human commonsense about their word choice in various contexts \cite{TransOMCS}. Nonetheless, the plausibility defined in commonsense knowledge merely indicates whether the statement makes sense and does not correspond exactly to SP. For example, the statements ``running requires shoes'' and ``running requires running shoes'' both make sense, so they have the same plausibility in ConceptNet, but humans prefer to select ``requires running shoes'' as the object of ``running''. In recent works~\cite{romero2019commonsense,chalier2020joint}, salience is proposed as a reflection of whether humans would spontaneously associate predicate and object with the given subject.

Dice~\citep{chalier2020joint} models four different dimensions of CSK statements: plausibility, typicality, remarkability and salience. Salience is expressed by the following logical constraints:
\begin{equation}
\begin{small}
\textrm{Typical}(s,p)\!\land\!\textrm{Remarkable}(s,p)\!\to\!\textrm{Salient}(s,p) 
\end{small}
\end{equation}
which means that if tuple $(s,p)$ is typical and $(s,p)$ is remarkable then $(s,p)$ is salient. Here $\textrm{Typical}$ reflects whether the property holds for most instances of the concept, and $\textrm{Remarkable}$ reflects the specific property of a concept that distinguishes it from highly related concepts. The definition of $\textrm{Remarkable}$ is based on the taxonomy of the concept. When the taxonomy is absent, highly related concepts are difficult to find.

CausalNet~\citep{luo2016commonsense} aims at extracting causal relations between terms. They propose necessity and sufficiency to model the causality strength between terms.
Necessity represents that the cause must be present in order for effect to take place. Sufficiency represents that cause is all it takes to bring about the effect. Motivated by the observation that CausalNet has shown effectiveness in SP tasks, we propose decomposing the salience metrics into sufficiency and necessity.

Sufficiency and necessity are also leveraged as auxiliary criteria in the annotation process to determine salience. We conduct regression experiments based on the annotation results, using sufficiency and necessity as independent variables and salience as dependent variables. The results show that the R-squad score is 0.566, and Prob(F) value is 0.00. This implies that sufficiency and necessity do have some validity in fitting salience. 
The coefficient of sufficiency is 0.62, and Prob(t) is 0.00. The coefficient on necessity is 0.39, and Prob(t) is 0.00. It means that necessity and sufficiency both have a statistically significant relationship with salience in the model.

\section{Knowledge Collection}
\label{sec:collection}
The steps of the commonsense knowledge collection are: 
\begin{enumerate}
    \item We match the concepts (including audience, event, category) in the pre-constructed knowledge base with those in free text (including shopping guidance and product title). 
    \item We ask annotators to annotate whether the N-tuple is plausible in the texts. 
    \item After annotation, we adopt the relation classifier model, Match The Blank (MTB)~\shortcite{soares2019matching}, which is based on task agnostic relation representations and tuned on supervised relation extraction dataset to learn the predicate between subject and object.
\end{enumerate}
Based on the idea of active learning, the whole process of mining commonsense knowledge is iterative. The relation we extract are only e-commerce related, including ``requires'', ``complementary'' and ``capable of''.

\section{Lexical Cues}
\label{sec:cues}
The application $\alpha_k$ of a cue $k$ is the number of instances that cue $k$ occur with one label but not with any others:
\begin{equation}
\alpha_k =\sum_{i=1}^n \mathbb I [\exists j,k \in \mathbb T_j^{(i)} \wedge k \notin  \mathbb T_{\neg j}^{(i)}]
\end{equation}% 
where $ \mathbb I $ is the indicator function (outputs one if the input is true; else 0). $ \mathbb T $ is the set of tokens, for instance, $i$ with label $j$. $n$ is the number of instances.

Coverage $\xi_k = \alpha_k/n$ of a cue $k$ is the proportion of applicable cases over all instances.

\section{Models Implementation Details}
\label{sec:models}
\paragraph{BERTSAGE.}The idea of BERTSAGE is to leverage the neighbor information of nodes through a graph neural network layer for their final embedding. The embeddings of $s,p,o$ are encoded by BERT separately. The inputs of BERT are the string of $s,p,o$ respectively, and the outputs are the embeddings of the [CLS] tokens in BERT's output layer denoted as $E_s, E_p, E_o$. For the embedding of node $s$, the final embedding is $\hat{E}_{s}=[E_s, \sum_{v \in \EuScript{N}(s)} E_v/|\EuScript{N}(s)|]$, where $\EuScript{N}(s)$ is the neighbor function that returns the neighbors of node $s$. Then salience score is predicted from the concatenated features $[\hat{E}_s; E_p; \hat{E}_o]$ by a dense layer.

\paragraph{KG-BERT.}KG-BERT(a)~\cite{yao2019kg} takes texts of $s,p,o$ as input of bidirectional encoder such as BERT and computes the scoring function of the triple with a language model. In specific, the input of model is the concatenation of $s,p,o$, as [CLS] $s$ [SEP] $p$ [SEP] $o$ [SEP]. The final hidden state $C$ corresponding to [CLS] and the classification layer weights $W$ are used to calculate the triple score.

\paragraph{StAR.}As KG-BERT doesn't utilize structured knowledge in the textual encoder, StAR~\cite{10.1145/3442381.3450043} divides the $s,p,o$ triple into $s,p$ and $o$ and encodes both parts by a Siamese-style textual encoder. Specifically, the two inputs are [CLS] $s$ [SEP] $p$ [SEP] and [CLS] $o$ [SEP] and after transformer encoder the outputs are the embeddings of the [CLS] tokens in BERT's output layer, namely $u$ and $v$. The final embeddings is $c=[u;u\times v;u-v;v]$. Then $c$ is fed into a neural binary classifier to get a score.

\paragraph{GenKGC.} GenKGC~\cite{xie2022discrimination} converts knowledge graph completion to sequence-to-sequence generation task with the pre-trained language model. Here, we utilize the generative model BART~\citep{BART}. The same input as KG-BERT is fed into the encoder and decoder of BART, and the representation from the final output token [SEP] from the decoder is used for classification.

\paragraph{PKGC.} PKGC~\cite{2022PKGC} is based on the idea of prompt tuning. It converts each triple into a sentence through a hard template and adds soft prompts in several fixed positions, then fed into PLMs for classification.

\section{Comparison Experiment with Dice}
\label{sec:experiment}
Dice~\cite{chalier2020joint} uses three CSK collections for evaluating the provided scores, i.e., ConceptNet, TupleKB, and Quasimodo. To obtain labeled data for hyper-parameter tuning and to serve as ground truth for evaluation, they sampled 200 subjects, each with two properties from each of the CSK collections, and asked annotators for pair-wise preference for each of the three dimensions using a 5-point Likert scale. 
 
Noted that they published the 3 CSK collections enriched with scores of four CSK dimensions (plausible, typical, remarkable, salient), we train our model PMI-tuning (original) on them. 
As the definition of remarkability and typicality is similar to necessity and sufficiency, we optimize necessity and sufficiency factors by the scores of remarkability and typicality, respectively.
 
Then we compare the results on the pair-wise preference (ppref) test set.  We adopt the same metrics of precision as Dice in the dimension of salience. The precision scores of Dice and baseline are provided by Dice. The baseline results come from the ranks of confidence scores from the original CSK collections. We observe that the best performance of the model is acquired in the first training epoch, and then the model gradually overfits the training set.

\section{Analysis}
\label{sec:Ablation}

\paragraph{Effect of the Encoders.}We evaluate different PLMs as encoders. Specifically, we choose RoFormer~\citep{su2021roformer}, RoBERTa~\citep{liu2019roberta}, BERT-wwm~\citep{cui2021pre} and MacBERT~\citep{cui2020revisiting}. 
RoFormer uses Rotary Position Embedding(RoPE) to encode positional information in transformer-based language models. BERT-wwm introduces the whole word masking (wwm) strategy to improve the performance of BERT. MacBERT proposes a new masking strategy called MLM as a correction (Mac) to pre-train PLMs based on RoBERTa. For a fair comparison, we choose the base version for every PLM. The experimental results are shown in Table \ref{tab:ablationencode}. As we can see from the table, BERT-wwm achieves the best performance. The possible reason is that the whole word mask strategy can compute more accurate word probability than the token mask strategy.

\begin{table}
\centering
\setlength{\tabcolsep}{4mm}{
\begin{tabular}{c|cccc}
\hline
\textbf{Encoder} &\textbf{F1}& \textbf{Acc.}  & \textbf{AUC} \\
\hline
RoFormer  & 60.1 & 59.8 & 70.5  \\
RoBERTa & 62.5 & 60.6 & 72.3  \\
BERT-wwm  & {\bf63.0} & {\bf62.1} & {\bf73.5} \\
MacBERT  & 62.0 & {\bf62.1} & 73.2 \\
\hline
\end{tabular}}
\caption{Test results on the simplified dataset for various encoders.}
\label{tab:ablationencode}
\end{table}

\paragraph{Effect of the Factors.}The value of $\lambda$ is set to 0 means that only the necessary factor is considered, and it is set to 1 means that only the sufficient factor is considered. We set $\lambda \in \{0,0.3,0.5,0.7,1\}$ to investigate the importance of necessity and sufficiency. The results are shown in Table \ref{tab:ablation}. 
When computing salience using a large language model, the pre-training corpus is largely unbiased. Therefore $\lambda=0.5$ gives the best result, suggesting that necessity is equally important as sufficiency.
We can see that $\lambda \in \{0.7,1\}$ model performance degradation is less than $\lambda \in \{0, 0.3\}$. It indicates that salience may be more related to sufficiency. 

\begin{table}
\centering
\setlength{\tabcolsep}{5mm}{
\begin{tabular}{c|cccc}
\hline
\textbf{$\lambda$} &\textbf{F1}& \textbf{Acc.}  & \textbf{AUC}   \\
\hline
0 & 62.0 & 60.3   & 70.1   \\
0.3 & 62.2 & 61.2   & 72.1    \\
0.5  & \bf{62.6} & \bf{63.3} & \bf{72.9}   \\
0.7 & 61.5 & 62.4   & 72.8      \\
1  & 61.1 & 61.0   & 72.7      \\
\hline
\end{tabular}}
\caption{Test results on the simplified dataset for different $\lambda$ variants.}
\label{tab:ablation}
\end{table}

\section{Application}
\label{sec:appendix}

To verify the correlation between the salience of commonsense triple and Click-Through Rate (CTR), we collect user click-through data from search logs in e-commerce. The query-product pair is selected when the query is the subject of a triple, the product title contains the object of this triple, and the predicate of this triple is ``requires''. Each triple corresponds to multiple query-product pairs, so the CTR of the triple is the average of the corresponding query-product CTRs. The influence of position bias of CTR is eliminated by the approach used in~\citep{yao2021learning}. Then we found a moderate positive correlation between CTR and the predicted salience score of triples, with a spearman coefficient of 0.42, which indicates the potential of salience to play a greater role in search.

Furthermore, salient common sense can be used effectively in many applications in e-commerce, such as product tagging. We need to tag products with multiple related concepts to better manage online products, for example, tagging cameras and suitcases with the concept of `travel.' 
The results of the tagging will be used to better support product search and recommendation. 
However, with 100 billion products online in e-commerce, it is not easy to use an end-to-end model to tag products to related concepts efficiently. 
In comparison, it is a simple and effective way to leverage commonsense knowledge to retrieve products and tag them. In this scenario, the salient triple ``running requires running shoes'' makes it easier to find products related to concepts than the plausible but non-salient triple ``running requires shoes''. The tagging system empowered by salient commonsense is ongoing at the time of writing. With the constraints of product category, the tagging system has achieved an accuracy of over 75\%. Compared to the commonsense triples, which have not been distinguished from salience, the accuracy is improved by approximately 20\%, saving a significant amount of computational cost for end-to-end models.

\end{document}